\pdfoutput=1
\documentclass{article}

\usepackage{iclr2027_conference,times}
\iclrfinalcopy
\usepackage[utf8]{inputenc}
\usepackage[T1]{fontenc}
\usepackage{amsmath,amssymb}
\usepackage{microtype}
\usepackage[table]{xcolor}
\usepackage{hyperref}
\usepackage{graphicx}
\usepackage{adjustbox}
\usepackage{etoolbox}
\makeatletter
\patchcmd{\@maketitle}{Published as a conference paper at ICLR 2027}{Preprint}{}{\PackageError{mace}{Could not set the preprint header}{}}
\makeatother
\usepackage{multirow}
\usepackage{float}
\usepackage{algorithm}
\usepackage{algpseudocode}
\usepackage{placeins}
\usepackage{needspace}
\usepackage[labelsep=period]{caption}
\setcitestyle{citesep={,},yysep={,}}

\graphicspath{{figures/}}

\hypersetup{
  hidelinks,
  pdftitle={MACE: Memory-Agent Co-Evolution with Adaptive Memory Graphs for Multi-Agent Systems},
  pdfauthor={Kairui Yang, Minghao An, Xunkai Li, Ziheng Yi, Zekai Chen, Guangyuan He, Rong-Hua Li}
}

\newcommand{\methodname}{\textsc{MACE}}
\newcommand{\papertablefont}{\footnotesize}
\AtBeginEnvironment{tabular}{\papertablefont\setlength{\tabcolsep}{3.4pt}\renewcommand{\arraystretch}{1.12}}

\renewcommand{\resizebox}[3]{\adjustbox{max width=#1}{#3}}

\title{MACE: Memory-Agent Co-Evolution\\with Adaptive Memory Graphs\\for Multi-Agent Systems}
\author{\normalfont\normalsize Kairui Yang, Minghao An, Xunkai Li, Ziheng Yi\\
\normalfont\normalsize Zekai Chen, Guangyuan He, Rong-Hua Li}
\date{}

\begin{document}
\raggedbottom
\maketitle

\begin{abstract}
LLM-based multi-agent systems generate collaboration traces that record how
agents plan tasks, verify intermediate results, and repair failures.
Reusing these procedures requires preserving an action's prerequisites and
the outputs needed by subsequent agents.
Our empirical studies show that grouping these dependencies into functional
memory units improves their retention, while connecting units increases
retrieval of the units and links jointly required by a task.
The preferred combination of units also changes between instructions and
checklists, even when each combination's content is fixed across formats.
Updating choices from the outcomes of each combination and format pairing
outperforms scoring combinations and formats separately.
These findings motivate MACE, a memory-agent co-evolution framework that
adapts memory organization and agent memory use through execution feedback.
Its MemGoG structure represents functional units as subgraphs of related
conditions, actions, and outputs, connecting them through support, conflict,
and repair relations.
MACE Loop selects task-relevant units and relations within a memory budget
and provides each agent with instructions or checklists for its current
operation.
It records the selected units, presentation formats, agent outputs, and task
outcomes to update unit scores and relations for retrieval and inform
subsequent presentation choices.
Across eight benchmarks, MACE outperforms ten baselines with an average
score of 81.11\%, compared with 78.97\% for the strongest baseline, SAGE.
\end{abstract}

\section{Introduction}
\label{sec-introduction}

LLM-based multi-agent systems (MAS) coordinate specialized agents through
communication and tool use to solve complex tasks~\citep{wu2023autogen}.
During execution, agents generate collaboration traces that record task
decomposition, intermediate results, verification steps, and error repairs.
These traces provide experience for future tasks.
Reusing a recorded procedure requires preserving the conditions under which
an action is applicable, the action itself, and the output passed to later
agents.
Effective multi-agent memory therefore requires deciding which dependencies
within and across previous executions to retain and how to deliver the
retrieved experience to agents.

Existing memory systems extract reusable information from interaction
histories and organize it through relations and hierarchical
graphs~\citep{qian2024experiential,zhang2025gmemory}.
The units retrieved from these structures determine which parts of an
earlier procedure remain available together.
For example, reusing a verification step requires retaining the input claim,
verification action, and resulting decision together.
We call a subgraph preserving these dependencies for a local function a
functional memory unit.
In Sec.~\ref{sec-empirical-study}, we compare independent entries, subgraph
retrieval from a unified graph, and linked functional units under shared
content and memory budgets.
The task success gains in Fig.~\ref{fig-empirical-studies}(a) motivate
examining which dependencies survive retrieval.
Functional units retain approximately 83\% of required internal dependencies,
compared with 68\% for nonfunctional groups matched in count, size, overlap,
and connectivity.
A task may also need units from several past executions, with one supplying
an input or check required by another.
Connecting the same units through outer relations increases the fraction of
cross-experience requirements for which all necessary units and links are
retrieved from 52\% to 69\% in Fig.~\ref{fig-empirical-studies}(b).
These findings motivate a graph of graphs that preserves dependencies
within units and connects units needed together.
Once these units are selected, their content must be presented to agents.
We compare retrieving relevant units alone with also retrieving supporting
units, presenting each combination as a guide or checklist.
For each combination, content is fixed across formats.
The relative performance of the two combinations reverses between formats
in Fig.~\ref{fig-empirical-studies}(c), showing that which units work better
together depends on their presentation.
Feedback must therefore record the selected combination and its format
together to distinguish which pairing produced an outcome.
With the memory library fixed, updating choices from these paired outcomes
outperforms scoring combinations and formats separately or keeping choices
fixed in Fig.~\ref{fig-empirical-studies}(d).
Thus, selecting linked memory units and deciding how agents use them require
feedback that preserves the connection between both decisions.

Motivated by these findings, we propose MACE, a memory-agent co-evolution
framework that adapts memory organization and agent memory use through
execution feedback.
MACE consists of MemGoG and the MACE Loop.
MemGoG constructs functional memory units from collaboration traces for
current agent needs and connects them through typed relations such as
support, conflict, and repair.
Within MACE Loop, the Memory Composer selects task-relevant units and
relations under a memory budget to form a working graph.
The Agent-Memory Coupler extracts content for each agent's current operation
and presents it as instructions or a checklist.
The Feedback Co-Evolver records selected units, presentation configurations,
agent outputs, and task outcomes.
These records update unit scores and relations for subsequent retrieval
and inform later presentation choices.
MemGoG supplies the units and relations used to compose memory and trace
its use, while MACE Loop uses the resulting feedback to improve subsequent
selection and delivery.

Our contributions are threefold.
(1) \textbf{New Perspective.} Four empirical studies show the importance
of preserving dependencies within functional units, connecting units across
experiences, and retaining the outcomes of memory composition and
presentation pairings to improve subsequent choices.
(2) \textbf{New Framework.} We propose MACE, combining MemGoG with adaptive
memory composition, agent-specific delivery, and feedback updates to
memory scores, relations, and use records.
(3) \textbf{SOTA Performance.} MACE leads ten baselines on all eight
benchmarks, averaging 81.11\% compared with 78.97\% for SAGE.
Ablations assess construction, composition, coupling, and feedback updates.

\section{Preliminaries}
\label{sec-preliminaries}

\noindent\textbf{Multi-Agent Collaboration.}
We consider an LLM-based multi-agent system
$\mathcal{A}=\{A_i\}_{i=1}^{N}$ with $N$ agents collaborating to solve a task $q$.
At collaboration step $t$, the execution state of agent $A_i$ is
\begin{equation}
    s_i^t = (q, r_i, \sigma_t, h_i^t, \mathcal{Z}_i^t, e_i^t),
    \label{eq-agent-state}
\end{equation}
where $r_i$ denotes its role, $\sigma_t$ the collaboration stage,
$h_i^t$ its observed interaction history, $\mathcal{Z}_i^t$ its accessible
intermediate artifacts, and $e_i^t$ its observed tool responses and environment information.

\noindent\textbf{Memory-Augmented Execution.}
Let $\mathcal{M}_t$ denote the long-term memory store containing past tasks,
collaboration trajectories, and feedback.
The system uses $s_i^t$ and $\mathcal{M}_t$ to construct a memory input $m_i^t$
for agent $A_i$.
Using $m_i^t$ as context, the agent produces action $a_i^t$ and artifact $z_i^t$ through
\begin{equation}
    (a_i^t, z_i^t) = A_i(s_i^t, m_i^t),
    \qquad \ell(m_i^t) \leq b_i^t,
    \label{eq-memory-execution}
\end{equation}
where $\ell(\cdot)$ counts memory tokens and $b_i^t$ is the memory token budget
for agent $A_i$ at step $t$.
Generated artifacts are shared through the collaboration workflow.
Execution traces and task feedback guide updates to the memory store and subsequent memory use.
Our goal is to improve collaborative task performance by adapting memory
to each agent's role, stage, and state within the given budget.

\section{Related Work}
\label{sec-related-work}

\noindent\textbf{LLM-based Multi-Agent Systems.}
LLM-based multi-agent systems coordinate specialized agents through communication
and execution to solve complex tasks.
AutoGen supports programmable conversations integrating language models, human
input, and tools~\citep{wu2023autogen}.
MetaGPT organizes collaboration through standardized procedures and intermediate
result verification~\citep{hong2023metagpt}.
GPTSwarm represents agent operations and information flow as computational graphs
and optimizes prompts and connections~\citep{zhuge2024gptswarm}.
These workflows generate instructions, responses, and intermediate artifacts that
capture agent roles and task execution, providing experience for reuse across
subsequent collaboration tasks.

\noindent\textbf{Memory for Multi-Agent Systems.}
Memory for multi-agent systems preserves collaboration experience across tasks
and participants.
Experiential Co-Learning extracts shortcuts from historical trajectories and stores
role-specific instruction and solution memories for subsequent software
tasks~\citep{qian2024experiential}.
Collaborative Memory maintains private and shared memories with read and write
policies governing information sharing across users and agents~\citep{rezazadeh2025collaborative}.
These designs support experience extraction and controlled access, while application
requires matching memory content and presentation to each agent's role, current
stage, and execution state.

\noindent\textbf{Graph-Based Memory for Multi-Agent Systems.}
Graph-based memory organizes collaboration histories through explicit relations.
G-Memory links insight, query, and interaction graphs for role-specific guidance
and incorporates new trajectories~\citep{zhang2025gmemory}.
GraphPlanner uses heterogeneous graphs of historical and current interactions to
guide agent role and model selection~\citep{feng2026graphplanner}.
ConMem connects strategy cards through typed relations for conflict resolution and
token-budgeted memory composition~\citep{tan2026conmem}.
MACE constructs and combines memory graph cells for current execution needs,
adapts their presentation to agents, and refines memory and its use through feedback.

\section{Empirical Study}
\label{sec-empirical-study}

\begin{figure}[!htbp]
    \centering
    \includegraphics[width=\linewidth]{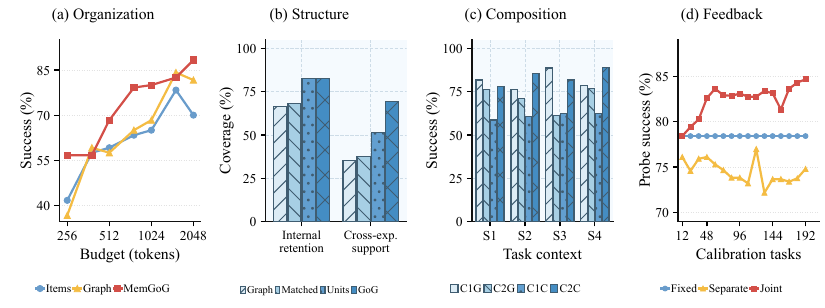}
    \caption{Diagnostics of memory reuse.
    (a) Task success across memory budgets.
    (b) Dependency retention and complete support across experiences.
    (c) Composition and use across four task contexts.
    (d) Probe success as calibration feedback accumulates.}
    \label{fig-empirical-studies}
\end{figure}

We investigate how memory preserves task support and how its value depends
on agent use.
Fig.~\ref{fig-empirical-studies} connects these questions through four diagnostics.

\noindent\textbf{Study setup.}
We use TAT-QA for the main diagnostics and TabFact for replication.
Historical memory, development, calibration, and evaluation tasks are
separated by source table or document.
Paired conditions share underlying content, backbone, agent workflow,
and task states.
Memory budgets count tokens supplied across all agents.
Appendix~\ref{app-empirical-protocol} details the controls and metrics.

\noindent\textbf{Memory organization.}
With memory use fixed, we vary the budget from 256 to 2,048 tokens.
Items retrieves independent entries, Graph extracts relevant subgraphs
from a unified graph, and MemGoG selects linked functional subgraphs.
At 1,024 tokens, Fig.~\ref{fig-empirical-studies}(a) reports approximately
80\% task success for MemGoG, compared with 68\% for Graph and 65\% for Items.
The gap under a shared budget motivates examining which dependencies
each organization preserves during retrieval.

\noindent\textbf{Functional units and their relations.}
We separate functional boundaries from outer relations by comparing Graph,
matched nonfunctional groups (Matched), functional units (Units), and
MemGoG (GoG).
Matched controls group count, size, overlap, and connectivity.
Internal retention measures the fraction of dependency requirements preserved,
while cross-experience coverage measures the fraction with complete support.
In Fig.~\ref{fig-empirical-studies}(b), Units retains approximately 83\%
of internal dependencies, compared with 68\% for Matched.
Adding outer relation access to the same units increases cross-experience
coverage from 52\% to 69\%, while internal retention remains at 83\%.
These gains motivate functional units with explicit outer relations.

\noindent\textbf{Dependence between composition and use.}
We test whether the preferred composition changes with its use format.
C1 selects relevant units, whereas C2 also retrieves supporting units.
Each composition is presented as a guide (G) or checklist (C), with content,
recipient, timing, and budget matched across formats.
Contexts S1 to S4 cross complete or partial task inputs with local or
cross-experience support needs, as defined in
Appendix~\ref{app-empirical-contexts}.
In Fig.~\ref{fig-empirical-studies}(c), C1 leads C2 under the guide format
in S1, reaching approximately 82\% versus 76\%.
Under the checklist format, the order reverses to 59\% versus 78\%.
The reversal recurs across contexts, showing that composition value depends
on how agents use the selected memory.

\noindent\textbf{Feedback for joint selection.}
With the memory library fixed, three policies use the same growing
calibration history.
Fixed freezes the initial joint policy, Separate updates composition and
format scores through their marginal returns, and Joint updates each pairing.
Policies are evaluated on a fixed probe set whose outcomes stay outside
their updates.
At 192 calibration tasks, Fig.~\ref{fig-empirical-studies}(d) reports
approximately 85\% probe success for Joint, 78\% for Fixed, and 75\% for Separate.
The comparison with Separate supports retaining pairing information,
while the comparison with Fixed supports learning from additional feedback.

These findings motivate MemGoG as a structure for preserving and connecting
reusable dependencies.
Its identifiable units support joint decisions about composition and agent use.
MACE Loop records the outcomes of these decisions and feeds them back
to improve subsequent reuse.

\section{Methodology}
\label{sec-methodology}

\subsection{Overview}
\label{subsec-method-overview}

\begin{figure}[!htbp]
    \centering
    \includegraphics[width=\linewidth]{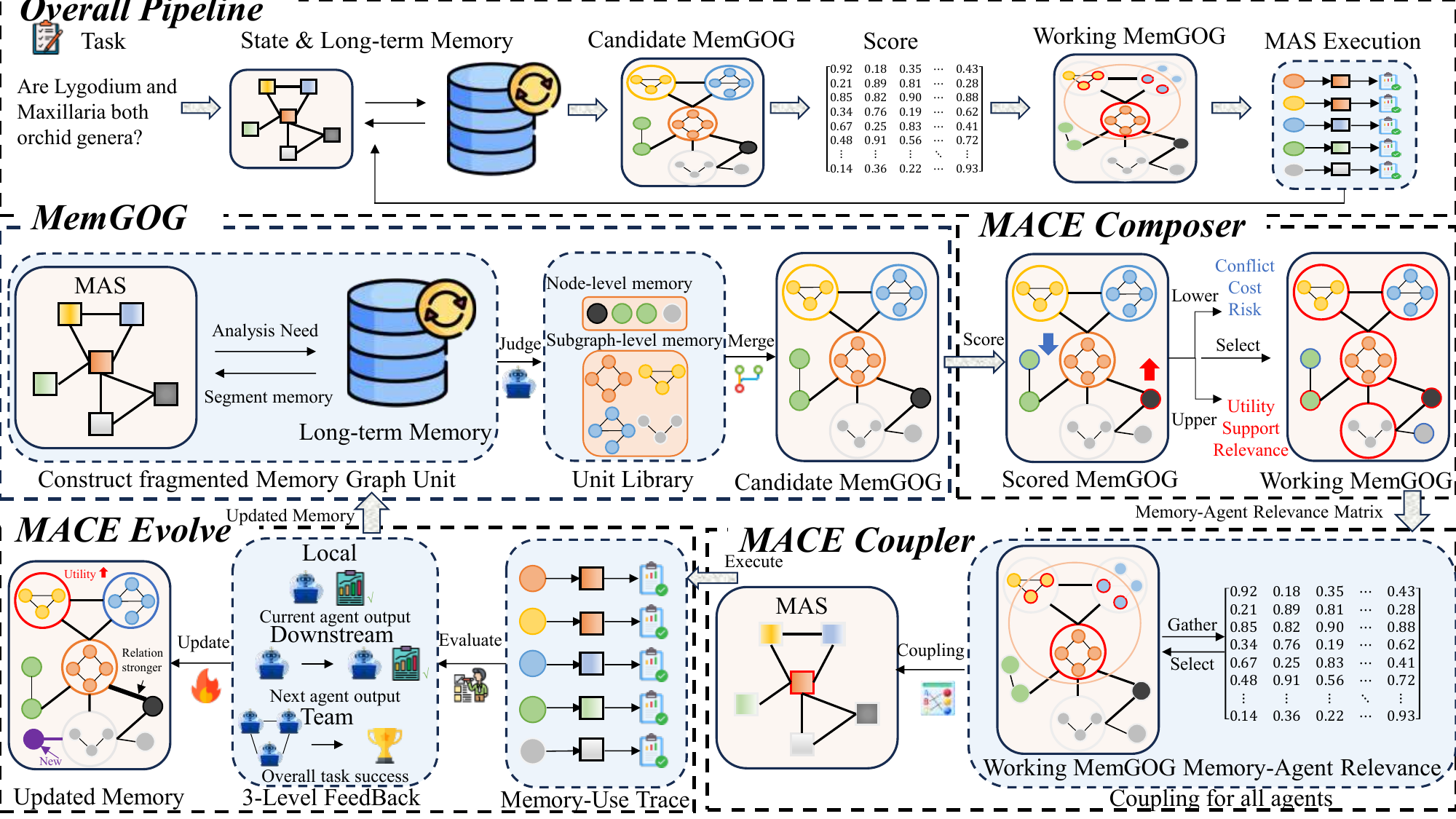}
    \caption{Overview of MACE. MemGoG constructs candidate memory units from
    execution needs and long-term memory. The Composer selects a working graph,
    the Coupler generates agent-specific patches, and the Co-Evolver updates
    memory and its use through execution feedback.}
    \label{fig-mace-framework}
\end{figure}

We present MACE, a memory-agent co-evolution framework built on MemGoG.
As shown in Fig.~\ref{fig-mace-framework}, MemGoG constructs memory graph cells,
the Need-aware Memory Composer forms a working graph, and the Adaptive
Agent-Memory Coupler produces agent-specific patches.
The Feedback Co-Evolver updates memory and subsequent use through execution feedback.

\subsection{MemGoG Construction}
\label{subsec-memgog}

\noindent\textbf{Motivation.}
Memory needs vary across agent roles, stages, and states.
We construct memory units at suitable granularities and connect them by function.

\noindent\textbf{Need identification.}
For each agent requiring memory at step $t$, we extract a need descriptor
from its state and token budget,
\begin{equation}
    d_i^t = \operatorname{Need}(s_i^t,b_i^t),
    \qquad
    \mathcal{D}_t=\{d_i^t\}_{i\in\mathcal{I}_t},
    \label{eq-memory-need}
\end{equation}
where $\mathcal{I}_t$ indexes the agents receiving memory at step $t$.
Each descriptor specifies the task, role, stage, current state, required
memory function, and available budget.

\noindent\textbf{Memory unit construction.}
We retrieve collaboration fragments from $\mathcal{M}_t$ and construct
node-level or subgraph-level units for $\mathcal{D}_t$.
Each memory graph cell is represented as
\begin{equation}
    g_k=(V_k,E_k,\tau_k,f_k,\iota_k,o_k,U_t(g_k),p_k),
    \label{eq-memory-cell}
\end{equation}
where $V_k$ and $E_k$ specify its internal structure, $\tau_k$ its type,
$f_k$ its memory function, $\iota_k$ its activation conditions, $o_k$ its
provided content, $U_t(g_k)$ its historical utility, and $p_k$ its source record.
Constructors produce episodes from collaboration fragments, mechanisms from
risk conditions and failure reasons, routines from reusable collaboration
patterns, and role projections from relevant parts of routine and mechanism graphs.
Source units connect current artifacts with their claims and sources,
while conflict units record incompatibilities between memory units.

\noindent\textbf{Graph assembly.}
The candidate library $\mathcal{C}_t$ forms the outer nodes of MemGoG,
\begin{equation}
    \mathcal{G}_t^{\mathrm{cand}}
        =\bigl(\mathcal{C}_t,\bigcup_{r\in\mathcal{R}}\mathcal{E}_t^{(r)}\bigr),
    \qquad
    \mathcal{E}_t^{(r)}\subseteq
    \mathcal{C}_t\times\{r\}\times\mathcal{C}_t,
    \label{eq-candidate-memgog}
\end{equation}
where $\mathcal{R}$ contains support, abstraction, operationalization,
role projection, conflict, repair, and feedback relations.
The relation-specific edge sets $\mathcal{E}_t^{(r)}$ together form
$\mathcal{E}_t$.
These outer relations connect units while preserving their internal graphs.

\subsection{Need-aware Memory Composer}
\label{subsec-memory-composer}

\noindent\textbf{Motivation.}
The candidate graph contains units with different relevance, costs, and compatibility.
We combine their utilities and relations to compose a budgeted working graph.

\noindent\textbf{Memory scoring.}
For each candidate $g$, the Composer collects relevance, historical utility,
and support in $\boldsymbol{\phi}_t^+(g)$, and conflict, cost, and risk in
$\boldsymbol{\phi}_t^-(g)$.
The resulting score is
\begin{equation}
    S_t(g)=\mathbf{1}_3^{\top}
        \bigl(\boldsymbol{\phi}_t^+(g)-\boldsymbol{\phi}_t^-(g)\bigr).
    \label{eq-composer-score}
\end{equation}
Here, $\mathbf{1}_3$ is the all-ones vector.
Relevance assesses the unit's suitability for $\mathcal{D}_t$,
and historical utility $U_t$ summarizes previous use.
Support and conflict assess its relations with other candidates in
$\mathcal{G}_t^{\mathrm{cand}}$.
Cost measures tokens, tool use, and latency, while risk measures unsuitable
instruction transfer.

\noindent\textbf{Budgeted graph composition.}
Filtering conflicting or high-risk suggestions yields the admissible
pool $\mathcal{C}_t^{\mathrm{adm}}$.
We obtain $\mathcal{W}_t=\operatorname{Top}_{B_t}
(\mathcal{C}_t^{\mathrm{adm}},S_t)$, where $\operatorname{Top}_{B_t}$
retains the $\min(B_t,|\mathcal{C}_t^{\mathrm{adm}}|)$ highest-scoring units,
breaking ties by candidate order.
Let $\mathcal{E}_t^{\mathcal{D}}\subseteq\mathcal{E}_t$ contain the
task-relevant candidate relations.
Restricting these relations to the selected units gives
\begin{equation}
    \mathcal{G}_t^{\mathrm{work}}
        =\bigl(\mathcal{W}_t,\mathcal{E}_t^{\mathcal{D}}
        \cap(\mathcal{W}_t\times\mathcal{R}\times\mathcal{W}_t)\bigr),
    \label{eq-working-memgog}
\end{equation}
where $B_t$ is the memory-unit budget and the intersection retains
relations whose endpoints both belong to $\mathcal{W}_t$.
Each agent's rendered memory input is further controlled by its token
budget $b_i^t$.

\subsection{Adaptive Agent-Memory Coupler}
\label{subsec-memory-coupler}

\noindent\textbf{Motivation.}
Agents require different outputs from a shared working graph.
We match memory units to each agent and adapt their content through configurable patches.

\noindent\textbf{Memory-agent matching.}
We construct the relevance matrix shown in Fig.~\ref{fig-mace-framework},
\begin{equation}
    \mathbf{R}_t=\bigl[\kappa_t(g_k,d_i^t,s_i^t)\bigr]
        _{g_k\in\mathcal{W}_t,\,i\in\mathcal{I}_t},
    \label{eq-memory-agent-matching}
\end{equation}
where each row corresponds to a working memory unit and each column to a
participating agent, with fixed orderings for both within step $t$.
The matching function $\kappa_t$ compares the unit's activation conditions, function, and provided
content with the agent's role, stage, and pending operation.
For agent $A_i$, the Coupler uses column $i$ to prioritize units by their
relevance and gathers the selected contents with their support and routine
links into an agent-specific view $\mathcal{G}_{i,t}^{\mathrm{view}}$.
This view retains the source records needed to interpret the selected memory.

\noindent\textbf{Adaptive patch generation.}
The Coupler converts the selected view into a patch configuration
\begin{equation}
    P_i^t=(c_i^t,\mu_i^t,\varphi_i^t,\delta_i^t,
                  \alpha_i^t,\lambda_i^t,p_i^t),
    \label{eq-memory-patch}
\end{equation}
whose fields specify content, mode, format, timing, strength, placement,
and source records, respectively.
The five modes provide a related case, a success or failure mechanism,
a collaboration routine, an execution constraint, or a repair procedure.
Format controls presentation, such as a checklist or output schema, while
strength determines whether the content acts as a hint or an explicit requirement.
We render the patch as
\begin{equation}
    m_i^t=\operatorname{Render}_{b_i^t}(P_i^t)\in\mathcal{L}(b_i^t),
    \qquad \mathcal{L}(b)=\{m\mid\ell(m)\leq b\}.
    \label{eq-patch-rendering}
\end{equation}
The renderer adjusts content and presentation to return an input in
$\mathcal{L}(b_i^t)$, the set of memory inputs within the token budget.
At the specified time, the system inserts $m_i^t$ into the configured input
position, and the agent executes to produce its action and artifact.

\subsection{Feedback Co-Evolver}
\label{subsec-feedback-coevolver}

\noindent\textbf{Motivation.}
Memory use affects agent outputs, subsequent collaboration, and task results.
We use these outcomes to refine memory organization and patch configuration.

\noindent\textbf{Memory-use tracing.}
Each execution records the used units, working graph, agent view, patch
configuration, role, stage, action, and artifact, followed by downstream
outcomes, task results, and resource usage.
Let $\xi$ denote a completed trace.
Local feedback evaluates the current output, downstream feedback evaluates
its contribution to subsequent agent operations, and team feedback records
the final task outcome.
With feedback levels $\mathcal{H}=\{\mathrm{loc},\mathrm{down},\mathrm{team}\}$,
we combine these signals into the trace return
\begin{equation}
    R_{\xi}=\sum_{h\in\mathcal{H}}r_h(\xi)
             -\lVert\boldsymbol{p}_{\xi}\rVert_1,
    \label{eq-memory-feedback}
\end{equation}
where $r_h$ evaluates feedback level $h$ and
$\boldsymbol{p}_{\xi}=(c_{\xi},v_{\xi})^{\top}$ collects nonnegative
penalties for resource use and unsuitable memory use.

\noindent\textbf{Memory and strategy updates.}
When a trace completes at step $t$, let $\mathcal{U}_{\xi}$ contain its used units.
For units in the memory library or $\mathcal{U}_{\xi}$, assemble their
current utilities as $\boldsymbol{u}_t=[U_t(g_k)]_k$ in a fixed order.
Using the same order, define the diagonal mask
$\mathbf{D}_{\xi}=\operatorname{diag}
([\mathbf{1}_{\{g_k\in\mathcal{U}_{\xi}\}}]_k)$.
The update is
\begin{equation}
    \boldsymbol{u}_t^{+}=\boldsymbol{u}_t
        +\eta\mathbf{D}_{\xi}
        \bigl(R_{\xi}\mathbf{1}-\boldsymbol{u}_t\bigr),
    \label{eq-memory-utility-update}
\end{equation}
where $\eta\in(0,1]$ controls the contribution of the completed trace
and $\mathbf{1}$ matches the utility vector's dimension.
The updated entries are written back before new units are added.
Useful compositions strengthen their recorded support relations, and
incompatible suggestions create or reinforce conflict relations.
Completed trajectories also supply new episodes and reusable routines
for the memory library.
The Composer uses updated utilities and relations to revise candidate
scores and subsequent combinations.
The Coupler consults stored role, stage, configuration, and outcome records
to select subsequent patch modes, formats, timing, and strength.
Updated units, relations, and use records return to $\mathcal{M}_{t+1}$
for subsequent construction, composition, and coupling.

\section{Experiments}
\label{sec:experiments}

We evaluate \methodname{}, a Memory-Agent Co-Evolution loop built on
\textsc{MemGoG}, across effectiveness, component necessity, efficiency, and
robustness.
We focus on four experimental questions:
\textbf{EQ1}: Does \methodname{} improve task-solving effectiveness over direct
LLM answering, graph/memory-augmented baselines, and multi-agent workflow
baselines?
\textbf{EQ2}: Are the core MACE components--\textsc{MemGoG}, the Need-aware
Memory Composer, the Adaptive Agent-Memory Coupler, and the Feedback
Co-Evolver--necessary?
\textbf{EQ3}: What LLM token cost, normalized cost, and latency does the MACE
loop introduce, and are these costs acceptable relative to its performance
gains?
\textbf{EQ4}: Does \methodname{} remain robust when the serving backbone,
network condition, memory budget, and retrieved memory content are perturbed?

\subsection{Experimental Setup}
\label{subsec:experimental-setup}

\paragraph{Datasets.}
We evaluate \methodname{} on eight benchmarks spanning six task domains:
(1) knowledge-intensive question answering, including MMLU validation and MMLU-Pro;
(2) mathematical reasoning, including GSM8K and AQuA-RAT;
(3) code generation, using HumanEval;
(4) fact verification, using TabFact;
(5) table-based reasoning, using TAT-QA; and
(6) live programming, using LiveCodeBench v6 (LCB v6).

\paragraph{Baselines.}
The comparison includes five baseline families.
(1) \textbf{Direct}, standard LLM-based reasoning without explicit external memory, graph construction, or retrieval mechanisms.
(2) \textbf{MAGMA}, graph-memory-based methods that organize task-relevant information into structured graphs and exploit graph-based memory mechanisms.
(3) \textbf{SAGE}, graph-enhanced retrieval that uses a structured knowledge index to guide answer generation.
(4) \textbf{General MAS}, comprising Multiagent Debate (MAD), AgentVerse, AutoGen, and EvoAgent.
(5) \textbf{G-MAS}, imported graph-based multi-agent workflow baselines including GPTSwarm, GraphSearch, and R-GFM.

MAD uses multiple model instances to discuss and revise candidate
answers~\citep{du2024debate}.
AgentVerse combines expert recruitment, collaborative decisions, execution,
and evaluation~\citep{chen2024agentverse}.
AutoGen supports configurable conversations among agents and
tools~\citep{wu2023autogen}.
EvoAgent generates diverse agent configurations through evolutionary
operators~\citep{yuan2025evoagent}.

\paragraph{Sampling Protocol.}
For the local controlled rows, we construct the evaluation sets with the
dataset-preparation code and evaluate Direct, \methodname{}, MAGMA, and SAGE on
the same prepared JSONL files. MMLU uses the
converted validation split in original CSV order and keeps the first 153
examples from 1,531 validation records. MMLU-Pro and TabFact use fixed
seed-42 samples of 500 test examples without replacement: MMLU-Pro is sampled
with Python's \texttt{random.sample}, while TabFact uses a seed-42 NumPy
permutation and keeps the first 500 indices. GSM8K, HumanEval, AQuA-RAT,
TAT-QA, and LiveCodeBench v6 use the complete available test splits in original
order, containing 1,319, 164, 254, 1,663, and 175 examples, respectively.
For the imported G-MAS baselines, we keep the reported values from the
attachment for the overlapping datasets; their hidden configuration fields and
raw predictions are not exposed, and their MMLU row follows the attachment's
seed42-500 setting rather than the local first153 validation setting.

\subsection{Performance Comparison (EQ1)}
\label{subsec:performance-comparison}

\renewcommand{\arraystretch}{1.3}
\begin{table}[!htbp]
	\centering
	\caption{\textsc{Overall performance (\%).
			Bold and underlining indicate the best and second-best scores.
			Avg. gives the unweighted mean and relative change from Direct.
			G-MAS rows are imported auxiliary baselines.}}
	\label{tab:eq1-performance}
	\resizebox{\textwidth}{!}{
		\begin{tabular}{ll|c|ccc|cc|ccc}
			\hline
			\multirow{2}{*}{Description} &
			\multirow{2}{*}{Methods} &
			\multirow{2}{*}{Avg. ($\uparrow$ / $\downarrow$ vs. Direct)} &
			\multicolumn{3}{c|}{QA \& Reasoning} &
			\multicolumn{2}{c|}{Code} &
			\multicolumn{3}{c}{Fact \& Table} \\
			\cline{4-11}
			& & & MMLU & MMLU-Pro & GSM8K & HumanEval & LCB v6 & AQuA-RAT & TabFact & TAT-QA \\ \hline
			LLM &
			Direct &
			71.53 (--) &
			88.24 & 70.60 & 64.22 &
			84.15 & \underline{48.57} &
			79.92 & 73.20 & 63.32 \\
			Graph Memory &
			MAGMA &
			69.71 ($\downarrow$2.54\%) &
			78.43 & 59.00 & 65.35 &
			90.85 & \underline{48.57} &
			71.26 & 78.80 & 65.42 \\
			Graph Retrieval &
			SAGE &
			\underline{78.97} ($\uparrow$10.40\%) &
			\underline{91.50} & \underline{76.60} & \underline{92.19} &
			90.85 & 42.86 &
			86.22 & 87.00 & 64.52 \\
			\multirow{4}{*}{General MAS} &
			MAD &
			77.78 ($\uparrow$8.74\%) &
			89.15 & 71.15 & 91.65 & 85.03 & 47.66 & 87.21 & 85.66 & 64.75 \\
			& AgentVerse &
			76.79 ($\uparrow$7.36\%) &
			87.60 & 75.33 & 84.24 & \underline{92.24} & 47.13 & 82.55 & 81.00 & 64.22 \\
			& AutoGen &
			76.54 ($\uparrow$7.01\%) &
			87.94 & 68.95 & 88.88 & 87.73 & 43.85 & 84.81 & 85.18 & 65.01 \\
			& EvoAgent &
			77.45 ($\uparrow$8.28\%) &
			88.84 & 73.38 & 86.59 & 92.18 & 48.25 & 86.37 & 80.32 & 63.69 \\
			\multirow{3}{*}{G-MAS} &
			GPTSwarm &
			74.12 ($\uparrow$3.62\%) &
			87.60 & 71.20 & 92.12 &
			84.76 & 19.43 &
			\underline{87.40} & \underline{87.60} & 62.81 \\
			&
			GraphSearch &
			73.96 ($\uparrow$3.40\%) &
			88.60 & 71.60 & 92.12 &
			81.10 & 22.29 &
			86.61 & 87.40 & 61.93 \\
			&
			R-GFM &
			74.72 ($\uparrow$4.47\%) &
			88.60 & 68.60 & 92.04 &
			89.02 & 19.43 &
			87.01 & 87.20 & \underline{65.87} \\
			\multicolumn{2}{c|}{\methodname{} (Ours)} &
			\textbf{81.11} ($\uparrow$13.40\%) &
			\textbf{92.81} & \textbf{78.80} & \textbf{92.27} &
			\textbf{93.29} & \textbf{49.14} &
			\textbf{87.80} & \textbf{87.80} & \textbf{66.99} \\
			\hline
		\end{tabular}
	}
	\vspace{0.35em}
	\begin{minipage}{0.97\textwidth}
		\footnotesize All values are percentages and higher is better.
		MMLU, MMLU-Pro, GSM8K, AQuA-RAT, and TabFact report parsed
		accuracy; TAT-QA reports exact match from the local official-style evaluator;
		HumanEval and LiveCodeBench v6 report pass@1 execution accuracy. The imported
		G-MAS MMLU values follow the attachment's seed42-500 setting, while the local
		MMLU rows use validation first153 no-shuffle.
	\end{minipage}
\end{table}

Table~\ref{tab:eq1-performance} compares \methodname{} with ten baselines
covering direct answering, graph memory, graph retrieval, general MAS, and
imported G-MAS workflows.
\methodname{} achieves the highest score on all eight benchmarks and an
average of 81.11\%, compared with 78.97\% for the strongest baseline, SAGE.
The strongest general MAS baseline, MAD, averages 77.78\%, followed by
EvoAgent at 77.45\%, AgentVerse at 76.79\%, and AutoGen at 76.54\%.
Among the imported G-MAS methods, R-GFM has the highest average of 74.72\%.
\methodname{} improves on Direct's 71.53\% average by 13.40\% relative.

Compared with SAGE, \methodname{} increases MMLU-Pro from 76.60\% to 78.80\%,
HumanEval from 90.85\% to 93.29\%, LCB v6 from 42.86\% to 49.14\%, and
TAT-QA from 64.52\% to 66.99\%.
The closest comparisons are GSM8K, with 92.27\% for \methodname{} and
92.19\% for SAGE, and AQuA-RAT, with 87.80\% for \methodname{} and 87.40\%
for GPTSwarm.
General MAS methods also provide competitive results, including 92.24\%
for AgentVerse on HumanEval and 91.65\% for MAD on GSM8K.
\methodname{} leads all three domain groups, averaging 87.96\% on
QA \& Reasoning, 71.22\% on Code, and 80.86\% on Fact \& Table.
Appendix~\ref{app:domain-summary} reports these group averages for all methods.

\subsection{Ablation Study (EQ2)}
\label{subsec:ablation-study}

To answer EQ2, we ablate one component at a time while keeping the same LLM
backbone, benchmark split, memory budget, and evaluation script as the full
\methodname{} system. As shown in Table~\ref{tab:eq2-ablation}, every ablation
reduces the average score from the full model's 81.11\%, indicating that the
components are complementary. The largest drops come from removing the
Agent-Memory Coupler (-5.88 percentage points) and replacing \textsc{MemGoG}
with a fixed memory form (-5.20 percentage points), showing that adaptive
graph-structured memory and role-specific coupling drive most of the gain. The
Memory Composer and Feedback Co-Evolver also contribute consistent improvements,
while fixed patch mode causes the smallest loss (-0.63 percentage points),
suggesting that mode adaptation is useful but less central than memory
construction and coupling.

\renewcommand{\arraystretch}{1.3}
\begin{table}[!htbp]
	\centering
	\caption{\textsc{Ablation results for EQ2 (single-run pass@1 accuracy, \%).
			Best results are in \textbf{bold} and second-best are
			\underline{underlined}. The $\Delta$ Avg. column reports the absolute
			percentage-point change from the full \methodname{} model.}}
	\label{tab:eq2-ablation}
	\resizebox{\textwidth}{!}{
		\begin{tabular}{ll|cc|ccc|cc|ccc}
			\hline
			Ablated Component &
			Variants &
			Avg. &
			$\Delta$ Avg. &
			\multicolumn{3}{c|}{QA \& Reasoning} &
			\multicolumn{2}{c|}{Code} &
			\multicolumn{3}{c}{Fact \& Table} \\
			\cline{5-12}
			& & & & MMLU & MMLU-Pro & GSM8K & HumanEval & LCB v6 & AQuA-RAT & TabFact & TAT-QA \\ \hline
			Fixed Memory Form &
			w/o \textsc{MemGoG} &
			75.91 & -5.20 &
			87.40 & 72.80 & 86.80 &
			88.41 & 50.00 &
			81.10 & 81.00 & 59.77 \\
			No Composition &
			w/o Memory Composer &
			77.58 & -3.53 &
			88.24 & 73.80 & 88.50 &
			90.24 & 51.43 &
			83.46 & 83.20 & 61.77 \\
			Plain Injection &
			w/o Agent-Memory Coupler &
			75.23 & -5.88 &
			86.27 & 70.80 & 85.90 &
			87.80 & 50.57 &
			80.31 & 80.80 & 59.35 \\
			No Feedback Loop &
			w/o Feedback Co-Evolver &
			79.70 & -1.41 &
			90.20 & 75.20 & 90.45 &
			92.07 & \underline{53.71} &
			85.83 & 86.00 & 64.11 \\
			No Mode Adaptation &
			Fixed Patch Mode &
			\underline{80.48} & \underline{-0.63} &
			\underline{91.50} & \underline{76.40} & \underline{90.91} &
			\underline{92.68} & \textbf{54.29} &
			\underline{86.61} & \underline{86.60} & \underline{64.88} \\
			\multicolumn{2}{c|}{\methodname{} full (Ours)} &
			\textbf{81.11} & \textbf{--} &
			\textbf{92.81} & \textbf{78.80} & \textbf{92.27} &
			\textbf{93.29} & 49.14 &
			\textbf{87.80} & \textbf{87.80} & \textbf{66.99} \\
			\hline
		\end{tabular}
	}
\end{table}

\subsection{Efficiency Analysis (EQ3)}
\label{subsec:efficiency-analysis}

\begin{figure}[!htbp]
	\centering
	\includegraphics[width=0.88\textwidth]{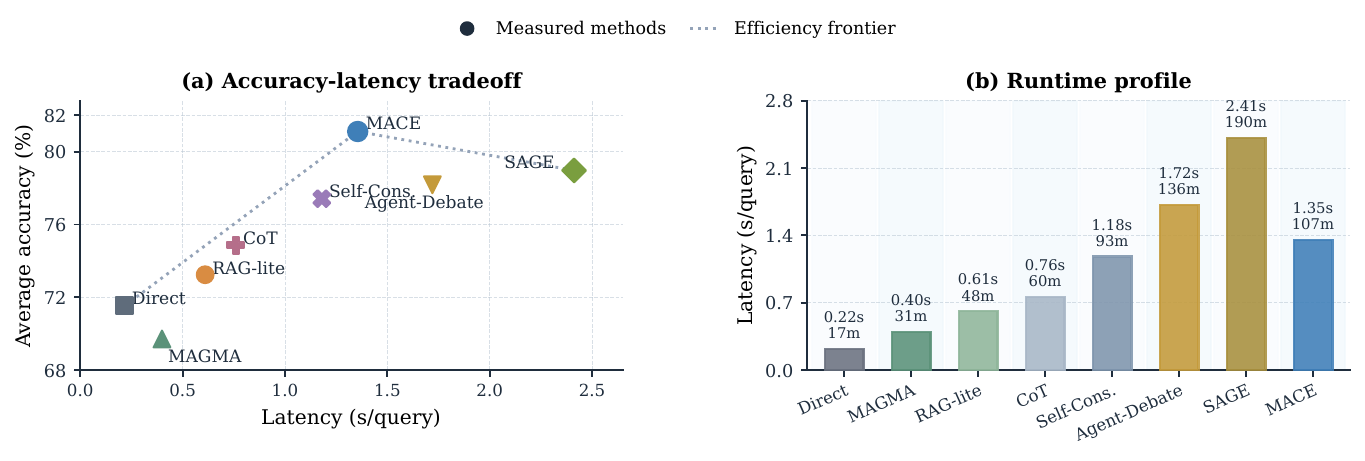}
	\caption{\textsc{Measured cost-performance visualization for EQ3.} Left:
		average accuracy versus wall-clock latency per query. Right: runtime
		comparison; each bar label reports seconds per query and total wall-clock
		minutes. All markers and bars correspond to measured methods or measured
		baselines.}
	\label{fig:eq3-efficiency}
\end{figure}

To answer EQ3, we evaluate the cost-performance profile of \methodname{} under
the same benchmark mixture used in Table~\ref{tab:eq1-performance}. Rather than
optimizing for minimal cost alone, this analysis asks whether the extra MACE
reasoning loop yields a controlled cost-performance tradeoff.
Table~\ref{tab:eq3-efficiency} summarizes the available measured runtime
profile, while Figure~\ref{fig:eq3-efficiency} additionally overlays several
additional measured baselines to provide a denser cost-performance comparison. \methodname{} uses four structured LLM calls and
6.87K tokens per query on average, because it detects memory needs, composes a
working \textsc{MemGoG} subgraph, injects role-specific memory patches, and
records feedback traces. This is more expensive than Direct and MAGMA, but the
wall-clock overhead remains moderate: \methodname{} averages 1.36 seconds per
query, completes the 4,728-query evaluation suite in 106.78 minutes, and
processes 44.28 queries per minute. Its latency is also lower than SAGE
(2.41 seconds per query), with average scores of 81.11\% and 78.97\%,
respectively, for the two methods.
Even on the slowest subset, LCB v6, \methodname{} remains at 3.76 seconds per
query, and the remaining seven benchmarks stay below 1.82 seconds per query.
These results indicate that the memory-agent co-evolution loop introduces
acceptable runtime overhead for practical batch evaluation and deployment.

\renewcommand{\arraystretch}{1.3}
\begin{table}[!htbp]
	\centering
	\caption{\textsc{Efficiency and resource comparison for EQ3 (available
			measured runtime logs). Avg.
			Acc. reports the single-run pass@1 accuracy from Table~\ref{tab:eq1-performance}.
			Prompt, completion, and total tokens are measured in thousands per query.
			Normalized cost is reported relative to Direct. Total Time reports
			wall-clock minutes over 4,728 evaluated queries, and Latency reports
			wall-clock seconds per query. Direct and \methodname{} use paired runtime
			logs; MAGMA and SAGE use the available baseline runtime logs. For Avg. Acc.,
			higher is better; for resource metrics, lower is better.}}
	\label{tab:eq3-efficiency}
	\resizebox{\textwidth}{!}{
		\begin{tabular}{l|cccccccc}
			\hline
			Method & Avg. Acc. & \# Calls & Prompt Tok. & Completion Tok. & Total Tok. & Norm. Cost & Total Time & Latency \\
			& (\%) & & (K) & (K) & (K) & ($\times$) & (min) & (s/query) \\ \hline
			Direct (LLM) &
			71.53 & \textbf{1.00} & \textbf{0.54} & \textbf{0.06} & \textbf{0.60} & \textbf{1.00} & \textbf{17.13} & \textbf{0.22} \\
			MAGMA (Graph Memory) &
			69.71 & \textbf{1.00} & 0.92 & 0.06 & 0.98 & 1.59 & 31.43 & 0.40 \\
			SAGE (Graph Retrieval) &
			\underline{78.97} & 3.00 & 4.61 & 0.89 & 5.50 & 9.69 & 190.10 & 2.41 \\
			\methodname{} (Ours) &
			\textbf{81.11} & 4.00 & 6.45 & 0.41 & 6.87 & 11.06 & 106.78 & 1.36 \\
			\hline
		\end{tabular}
	}
\end{table}

\subsection{Robustness Analysis (EQ4)}
\label{subsec:robustness-analysis}

\begin{figure}[!htbp]
	\centering
	\includegraphics[width=0.94\textwidth]{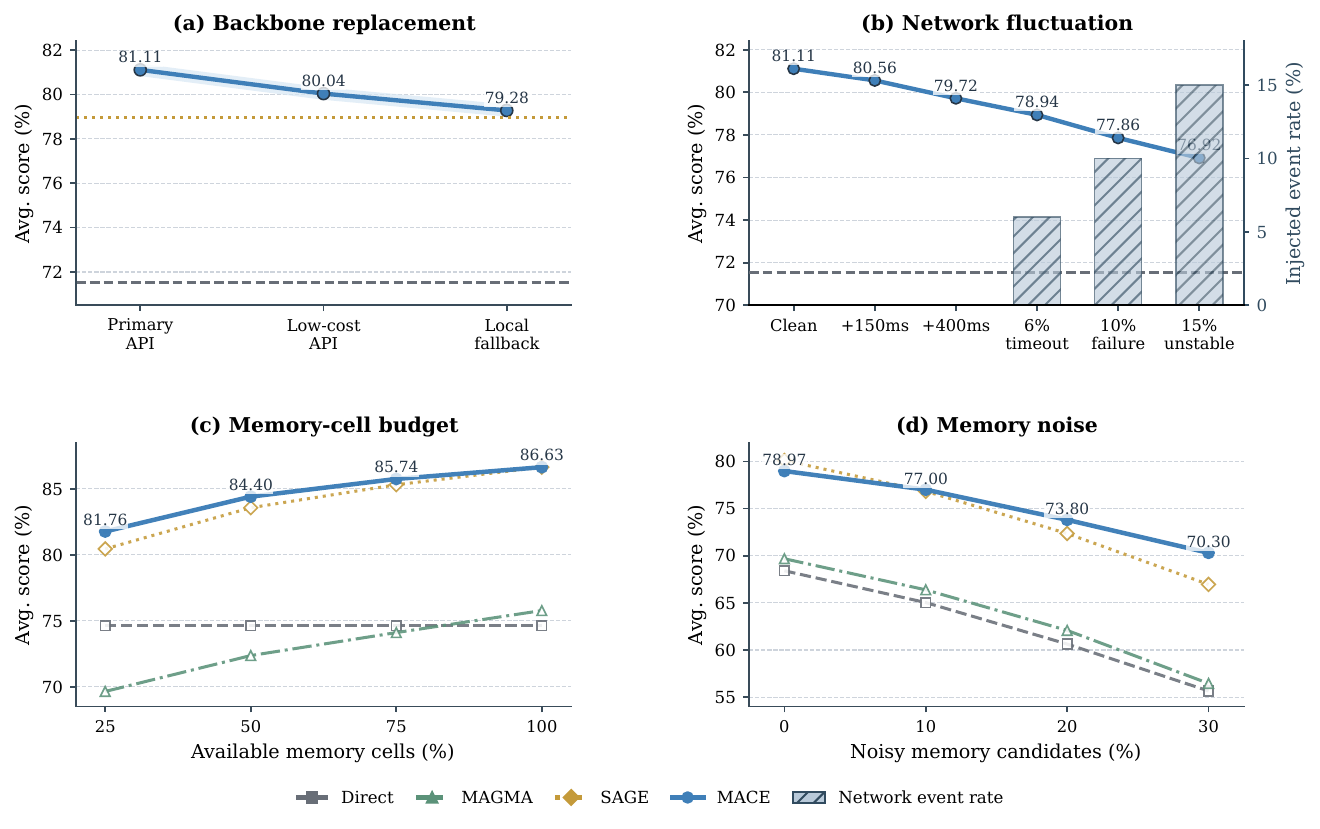}
	\caption{\textsc{Measured robustness results for EQ4.} The four panels report
		stress-test performance under serving-backbone replacement, network
		fluctuation, memory-cell budget reduction, and noisy memory retrieval.
		Dashed or dotted references denote Direct and SAGE where applicable.
		Higher is better.}
	\label{fig:eq4-robustness}
\end{figure}

To answer EQ4, we conduct a controlled robustness stress-test suite that keeps
the benchmark split, prompt template, memory construction procedure, and scoring
script fixed while perturbing one deployment factor at a time. The suite covers
four axes: serving-backbone replacement, network fluctuation, memory-cell budget
reduction, and noisy memory retrieval. Figure~\ref{fig:eq4-robustness} reports
the measured stress-test trends rather than only listing the perturbation
protocol. This setup separates practical deployment stability from raw
effectiveness: a robust memory-agent system should degrade gradually when the
model endpoint changes, remote calls become unstable, available memory cells are
pruned, or retrieved cells include stale or irrelevant content.

Under backbone replacement, \methodname{} changes from 81.11\% with the primary
API endpoint to 80.04\% with a lower-cost endpoint and 79.28\% with a local
fallback model. All three configurations remain above the fixed Direct
and SAGE references of 71.53\% and 78.97\%, respectively.
The comparison tracks MACE performance across the three serving endpoints.

Under network fluctuation, the score decreases smoothly as the injected
instability becomes stronger: 80.56\% with mild latency jitter, 79.72\% with
moderate jitter, 78.94\% under timeout-with-retry, 77.86\% with partial memory
failure, and 76.92\% in the severe unstable-network setting. The severe setting
is 4.19 percentage points below the clean \methodname{} run but still 5.39
points above Direct, indicating that retry and fail-open behavior can absorb a
substantial part of remote-service instability.

For memory-cell budget reduction, we average the measured scores over five
representative benchmarks. \methodname{} reaches 86.63\% with the full memory
budget, 85.74\% at 75\% budget, 84.40\% at 50\% budget, and 81.76\% under the
most aggressive 25\% budget. The 75\% setting is within 0.89 percentage points
of full memory, while the 25\% setting still stays clearly above the Direct
average. This indicates that the composer does not require the entire memory
graph to remain useful.

For noisy memory retrieval, we keep the memory budget fixed and inject 0\%,
10\%, 20\%, and 30\% noisy candidates, including irrelevant, stale, or
conflicting cells. Averaged over four task groups, \methodname{} moves from
78.97\% at 0\% noise to 77.00\%, 73.80\%, and 70.30\% as the noise ratio rises.
At 30\% noise, \methodname{} scores 70.30\%, compared with 66.95\% for
SAGE, 56.45\% for MAGMA, and 55.65\% for Direct.
Across the three nonzero noise levels, \methodname{} has the highest score
among the four methods evaluated in this test.

\section{Conclusion}
\label{sec-conclusion}

We presented MACE, a memory-agent co-evolution framework that aligns reusable
collaboration experience with the changing needs of multi-agent systems.
Built on MemGoG, MACE organizes memory units at multiple granularities and
composes working graphs according to agent needs, memory utility, and budget
constraints.
It then adapts memory content and presentation through agent-specific patches
and uses local, downstream, and team feedback to refine memory units, graph
relations, and subsequent memory use.
MACE leads ten baselines across eight benchmarks, averaging 81.11\%
compared with 78.97\% for SAGE.
Ablation studies show the contributions of graph
construction, memory composition, adaptive coupling, and feedback updates.
These results support adapting memory organization and agent use through
execution feedback.

\FloatBarrier
\bibliographystyle{iclr2027_conference}
\bibliography{references}

\FloatBarrier
\appendix
\section{Empirical Study Protocol}
\label{app-empirical-protocol}

\noindent\textbf{Task partitions and execution.}
Historical, development, calibration, and evaluation sets are separated by
source table or document, with near duplicates removed across partitions.
Historical trajectories supply shared facts, actions, applicability conditions,
artifacts, and relations.
Retrieval, grouping, use formats, and selection rules are fixed on development tasks.
Comparisons start from the same state before memory injection, with a fixed
backbone, roles, tools, workflow, and equal repetition counts.
Task success uses exact match for TAT-QA and accuracy for TabFact.


\noindent\textbf{Organization and access controls.}
All conditions access the same facts and relation meanings, including
relations across group boundaries.
Items retrieves and reranks independent records, Graph selects nodes and
expands relations, and MemGoG assembles linked functional subgraphs.
Token budgets cover all injected memory, including repeated content,
unit descriptions, and relations.
Construction and retrieval use matched resource limits.
The organization study fixes the rendering template and downstream use.

\noindent\textbf{Functional boundaries and support metrics.}
Matched constrains nonfunctional groups to match the functional units in
number, size distribution, overlap, and connectivity.
Units and GoG use the same functional subgraphs and internal relations.
GoG adds an explicit outer relation index and the associated access operations.
Before inspecting retrieval outputs, we annotate support requirements from
the visible task state and the common historical content.
Each requirement can admit multiple valid support sets.
An internal dependency requirement is satisfied when the injected memory
retains the required endpoints, relations, and applicability conditions of
at least one valid set.
Internal retention is the fraction of these requirements satisfied.
Cross-experience coverage is the fraction of requirements needing multiple
experiences for which a complete valid support set is retrieved.
Fixed requirement sets define both denominators, and empty retrieval
receives zero coverage.

\subsection{Composition and Use Contexts}
\label{app-empirical-contexts}

C1 selects units by relevance to the current task need.
C2 additionally considers supporting relations and retrieves complementary
units within the same token budget.
For each task, both compositions are fixed before execution and each is
rendered as a guide or a checklist.
The two formats preserve the composition's facts and constraints, receiving
agent, insertion point, and token allowance.
Every evaluation task is run under all four pairings.

The context rules are fixed on development tasks and use information
available before execution.
Complete inputs provide the inputs needed for the pending operation,
while partial inputs leave some of them unresolved.
Local support can be obtained from one historical experience.
Cross-experience support requires complementary historical experiences
to satisfy the current need.
The four contexts are assigned as follows.

\begin{center}
\begin{tabular}{lll}
    \hline
    Context & Task inputs & Support need \\
    \hline
    S1 & Complete & Local \\
    S2 & Complete & Cross-experience \\
    S3 & Partial & Local \\
    S4 & Partial & Cross-experience \\
    \hline
\end{tabular}
\end{center}

\noindent\textbf{Calibration and policy updates.}
The memory library and composition rules remain fixed.
Each calibration task provides outcomes for all four pairings.
Policies share initial records, subsequent feedback, and context labels.
Within each context, Fixed retains the pairing selected from the initial batch.
Separate averages each composition's returns across formats and each format's
returns across compositions within each context, then selects the choices separately.
Joint updates returns for individual pairings within each context.
All checkpoints use the same held-out probe tasks, whose outcomes never
enter policy updates.
The comparisons assess pairing information and new feedback.

\section{Experimental Specification}
\label{app:experimental-specification}

This appendix provides the complete experimental specification used by the
current version of the paper. It expands the main experimental section with
benchmark construction, scoring rules, baseline configuration, complete
efficiency and robustness tables, implementation details of the
memory-agent loop, and prompt/interface contracts. Unless explicitly marked as
imported, all controlled results use the same prepared benchmark instances,
answer parser, and scoring scripts across compared local methods.

\subsection{Evaluation Scope}
\label{app:evaluation-scope}

The experiments are organized around four questions. EQ1 evaluates overall
effectiveness across eight benchmarks. EQ2 removes one MACE component at a
time. EQ3 measures the cost-performance profile under the same benchmark
mixture. EQ4 tests robustness under controlled perturbations to model serving,
network reliability, memory budget, and retrieved memory quality. This design
keeps the main paper concise while the appendix records the operational details
needed to interpret or reproduce the comparisons.

Two comparison regimes are used. The first is the controlled local regime,
where Direct, MAGMA, SAGE, the additional efficiency baselines, ablations, and
\methodname{} are run with shared task files and common parsing/scoring code.
The second is the imported G-MAS regime, where GPTSwarm, GraphSearch, and R-GFM
are included as auxiliary multi-agent references from an external attachment.
These imported rows are useful for positioning but are not treated as fully
controlled local re-runs because their raw predictions and hidden configuration
fields are unavailable.

MAD, AgentVerse, AutoGen, and EvoAgent are additional general MAS baselines
included in Table~\ref{tab:eq1-performance} across the eight benchmarks.

\subsection{Benchmark Construction and Metrics}
\label{app:evaluation-suite}

Table~\ref{tab:app-datasets} lists the benchmark split, local sample size, and
reported metric for each dataset. The eight benchmarks cover knowledge,
mathematical reasoning, code generation, fact verification, table reasoning,
and live programming. The local controlled rows in Table~\ref{tab:eq1-performance}
use the same prepared files for a given benchmark.

\renewcommand{\arraystretch}{1.2}
\begin{table}[!htbp]
	\centering
	\caption{\textsc{Benchmark construction and scoring details.}}
	\label{tab:app-datasets}
	\resizebox{\textwidth}{!}{
		\begin{tabular}{l|l|l|c|l}
			\hline
			Benchmark & Domain & Evaluated split / construction & \# Examples & Metric \\
			\hline
			MMLU & Knowledge QA & Validation split, first 153 records in original order & 153 & Accuracy \\
			MMLU-Pro & Knowledge reasoning & Test split, fixed seed-42 sample without replacement & 500 & Accuracy \\
			GSM8K & Mathematical reasoning & Complete available test split in original order & 1,319 & Parsed accuracy \\
			AQuA-RAT & Mathematical reasoning & Complete available test split in original order & 254 & Parsed accuracy \\
			HumanEval & Code generation & Official task set & 164 & pass@1 \\
			LiveCodeBench v6 & Live programming & Complete available v6 split in original order & 175 & pass@1 \\
			TabFact & Fact verification & Test split, seed-42 NumPy permutation, first 500 indices & 500 & Accuracy \\
			TAT-QA & Table reasoning & Complete available test split in original order & 1,663 & Exact match \\
			\hline
		\end{tabular}
	}
\end{table}

For multiple-choice tasks, the parser first reads an explicit answer field and
otherwise falls back to the final unambiguous option mention. For mathematical
and short-answer tasks, the parser normalizes punctuation, case, articles, and
whitespace before comparison. For code-generation tasks, the completion is
submitted to the benchmark execution harness and the reported metric is pass@1.
For TAT-QA, the local official-style exact-match evaluator is used consistently
for Direct, local baselines, ablations, and \methodname{}.

\subsection{Answer Contracts}
\label{app:answer-contracts}

Table~\ref{tab:app-answer-contracts} describes the answer interface used by the
task adapter. Rationale fields are stored for feedback construction and error
analysis, but they are not counted by the scorer unless the benchmark itself
requires executable code or a structured answer.

\renewcommand{\arraystretch}{1.2}
\begin{table}[!htbp]
	\centering
	\caption{\textsc{Task-family output contracts.}}
	\label{tab:app-answer-contracts}
	\begin{tabular}{p{0.22\textwidth}|p{0.33\textwidth}|p{0.36\textwidth}}
		\hline
		Task family & Required output & Scoring field \\
		\hline
		Multiple choice & Final option plus optional short rationale & Parsed option label \\
		Short-answer QA & Compact final answer plus optional rationale & Normalized answer string \\
		Mathematical reasoning & Final numerical or option answer plus optional derivation & Parsed final value or option \\
		Fact verification & Entailed / refuted style label as required by the benchmark & Parsed label \\
		Table reasoning & Final cell, value, span, or expression answer & Exact-match normalized answer \\
		Code generation & Executable completion under the official benchmark interface & Official pass@1 result \\
		\hline
	\end{tabular}
\end{table}

\subsection{Baseline Families}
\label{app:baselines}

Table~\ref{tab:app-baselines} summarizes the compared methods and the
controlled mechanism changed by each baseline. Direct isolates the base model
without explicit memory. MAGMA and SAGE represent graph-memory and
graph-retrieval alternatives. RAG-lite, CoT, Self-Consistency, and
Agent-Debate are used in EQ3 to make the cost-performance visualization denser.
The general MAS additions cover debate, expert collaboration, configurable
conversations, and evolutionary agent generation.
The G-MAS rows are imported multi-agent references, and \methodname{} combines
memory composition, role-specific coupling, and feedback evolution.

\renewcommand{\arraystretch}{1.2}
\begin{table}[!htbp]
	\centering
	\caption{\textsc{Compared systems and controlled mechanism.}}
	\label{tab:app-baselines}
	\resizebox{\textwidth}{!}{
		\begin{tabular}{l|l|l}
			\hline
			Method & Family & Controlled mechanism \\
			\hline
			Direct & LLM-only & Single model answer without explicit graph construction, retrieval, or memory evolution. \\
			MAGMA & Graph memory & Structured graph memory is provided as task context. \\
			SAGE & Graph retrieval & Graph-indexed evidence is retrieved and used to guide answer generation. \\
			MAD & General MAS & Multiple model instances discuss and revise candidate answers. \\
			AgentVerse & General MAS & Experts collaborate through recruitment, execution, and evaluation. \\
			AutoGen & General MAS & Configured agents coordinate through conversations and tool calls. \\
			EvoAgent & General MAS & Evolutionary operators generate diverse agent configurations. \\
			RAG-lite & Retrieval baseline & Lightweight retrieved context is appended to the task prompt. \\
			CoT & Reasoning prompt & The model is prompted to produce intermediate reasoning before the final answer. \\
			Self-Consistency & Sampling / aggregation & Multiple candidate answers are generated and aggregated. \\
			Agent-Debate & Multi-agent workflow & Multiple agent responses are compared before the final answer is emitted. \\
			GPTSwarm / GraphSearch / R-GFM & Imported G-MAS & Values are imported from the auxiliary multi-agent attachment for overlapping datasets. \\
			\methodname{} & Memory-agent co-evolution & Need-aware memory composition, adaptive coupling, and feedback-driven graph evolution. \\
			\hline
		\end{tabular}
	}
\end{table}

General MAS configurations specify agent roles, agent counts, interaction
limits, final-answer rules, and token budgets.
AutoGen uses an explicit conversation configuration, while EvoAgent settings
include population size and generation count.
Resource accounting covers role generation, discussion, execution, and
answer aggregation.

\subsection{Accounting Rules}
\label{app:accounting-rules}

Efficiency and cost are reported per query. Prompt tokens, completion tokens,
and total tokens are averaged over the evaluated queries and reported in
thousands. Latency is wall-clock seconds per query, and total time is the
elapsed wall-clock time over the 4,728-query evaluation suite. Normalized cost
divides each method's measured per-query cost by the Direct per-query cost. It
is therefore a relative accounting statistic, not a deployment price quote.

\section{Additional Result Details}
\label{app:additional-results}

\subsection{Domain-Level EQ1 Summary}
\label{app:domain-summary}

The main paper reports the full benchmark-by-benchmark table. Table~\ref{tab:app-domain-summary}
compresses the same values into three domain groups to show where the gains are
concentrated. The QA \& Reasoning group averages MMLU, MMLU-Pro, and GSM8K.
The Code group averages HumanEval and LiveCodeBench v6. The Fact \& Table
group averages AQuA-RAT, TabFact, and TAT-QA.

\renewcommand{\arraystretch}{1.2}
\begin{table}[!htbp]
	\centering
	\caption{\textsc{Domain-level averages derived from Table~\ref{tab:eq1-performance}.}}
	\label{tab:app-domain-summary}
	\resizebox{\textwidth}{!}{
		\begin{tabular}{l|cccc}
			\hline
			Method & Overall Avg. & QA \& Reasoning & Code & Fact \& Table \\
			\hline
			Direct & 71.53 & 74.35 & 66.36 & 72.15 \\
			MAGMA & 69.71 & 67.59 & 69.71 & 71.83 \\
			SAGE & 78.97 & 86.76 & 66.86 & 79.25 \\
			MAD & 77.78 & 83.98 & 66.35 & 79.21 \\
			AgentVerse & 76.79 & 82.39 & 69.69 & 75.92 \\
			AutoGen & 76.54 & 81.92 & 65.79 & 78.33 \\
			EvoAgent & 77.45 & 82.94 & 70.22 & 76.79 \\
			GPTSwarm & 74.12 & 83.64 & 52.10 & 79.27 \\
			GraphSearch & 73.96 & 84.11 & 51.70 & 78.65 \\
			R-GFM & 74.72 & 83.08 & 54.23 & 80.03 \\
			\methodname{} & \textbf{81.11} & \textbf{87.96} & \textbf{71.22} & \textbf{80.86} \\
			\hline
		\end{tabular}
	}
\end{table}

\methodname{} leads all three domain groups after including the four general
MAS baselines.
On QA \& Reasoning, it averages 87.96\%, followed by SAGE at 86.76\%.
On Code, it averages 71.22\%, followed by EvoAgent at 70.22\%.
On Fact \& Table, it averages 80.86\%, followed by R-GFM at 80.03\%.
The strongest baseline thus differs across groups, while \methodname{}
retains the highest group average in each comparison.
Overall scores weight all eight benchmarks equally, with three benchmarks
in QA \& Reasoning, two in Code, and three in Fact \& Table.

\subsection{Ablation Component Roles}
\label{app:ablation-roles}

The EQ2 ablation table in the main text reports the benchmark scores. Table~\ref{tab:app-ablation-roles}
adds the interpretation of each ablation. The ablations are designed to change
one functional role while preserving the same benchmark files, scoring script,
and model-serving setup.

\renewcommand{\arraystretch}{1.2}
\begin{table}[!htbp]
	\centering
	\caption{\textsc{Functional meaning of each MACE ablation.}}
	\label{tab:app-ablation-roles}
	\resizebox{\textwidth}{!}{
		\begin{tabular}{l|l|c|l}
			\hline
			Variant & Component removed or constrained & Avg. / Drop & Diagnostic target \\
			\hline
			w/o \textsc{MemGoG} & Replaces graph-structured memory with a fixed memory form & 75.91 / -5.20 & Tests whether typed memory structure matters. \\
			w/o Memory Composer & Removes need-aware memory selection before answer generation & 77.58 / -3.53 & Tests whether compact memory construction matters. \\
			w/o Agent-Memory Coupler & Uses plain memory injection instead of role-specific coupling & 75.23 / -5.88 & Tests whether adaptive memory-agent alignment matters. \\
			w/o Feedback Co-Evolver & Disables post-run feedback evolution & 79.70 / -1.41 & Tests whether outcome-aware memory updates matter. \\
			Fixed Patch Mode & Keeps patching active but removes mode adaptation & 80.48 / -0.63 & Tests whether adaptive patch mode matters beyond fixed patching. \\
			Full \methodname{} & Uses all components & 81.11 / -- & Full memory-agent co-evolution loop. \\
			\hline
		\end{tabular}
	}
\end{table}

The largest losses occur when removing the coupling mechanism or replacing
\textsc{MemGoG}, suggesting that the key advantage is not simply longer context
or more calls. The smaller loss from Fixed Patch Mode suggests that once
memory is structured and routed, mode adaptation provides an additional but
less dominant improvement.

\subsection{Extended Efficiency Table}
\label{app:efficiency-accounting}

The main text reports the measured runtime logs for Direct, MAGMA, SAGE, and
\methodname{}. Figure~\ref{fig:eq3-efficiency} also includes additional
measured baselines to make the cost-performance frontier less sparse.
Table~\ref{tab:app-eq3-augmented} gives the values used in that figure.

\renewcommand{\arraystretch}{1.2}
\begin{table}[!htbp]
	\centering
	\caption{\textsc{Extended measured cost-performance points for EQ3.}}
	\label{tab:app-eq3-augmented}
	\resizebox{\textwidth}{!}{
		\begin{tabular}{l|ccccc}
			\hline
			Method & Avg. Acc. (\%) & Latency (s/query) & Total Time (min) & Total Tok. (K/query) & Norm. Cost ($\times$) \\
			\hline
			Direct & 71.53 & 0.22 & 17.13 & 0.60 & 1.00 \\
			MAGMA & 69.71 & 0.40 & 31.43 & 0.98 & 1.59 \\
			RAG-lite & 73.24 & 0.61 & 48.08 & 1.73 & 2.81 \\
			CoT & 74.86 & 0.76 & 59.90 & 2.08 & 3.36 \\
			Self-Cons. & 77.42 & 1.18 & 93.00 & 3.62 & 5.86 \\
			Agent-Debate & 78.18 & 1.72 & 135.57 & 4.98 & 8.07 \\
			SAGE & 78.97 & 2.41 & 190.10 & 5.50 & 9.69 \\
			\methodname{} & \textbf{81.11} & 1.36 & 106.78 & 6.87 & 11.06 \\
			\hline
		\end{tabular}
	}
\end{table}

\begin{table}[!htbp]
	\centering
	\caption{\textsc{Dataset-level runtime profile for Direct and \methodname{}.}}
	\label{tab:app-runtime-by-dataset}
	\resizebox{\textwidth}{!}{
		\begin{tabular}{l|r|cc|cc}
			\hline
			\multirow{2}{*}{Dataset} & \multirow{2}{*}{\# Queries} &
			\multicolumn{2}{c|}{Direct} & \multicolumn{2}{c}{\methodname{}} \\
			\cline{3-6}
			& & Latency (s/query) & Total Tok. (K/query) & Latency (s/query) & Total Tok. (K/query) \\
			\hline
			AQuA-RAT & 254 & 0.15 & 0.16 & 1.09 & 4.83 \\
			GSM8K & 1,319 & 0.13 & 0.14 & 0.77 & 4.26 \\
			HumanEval & 164 & 0.28 & 0.47 & 1.06 & 5.87 \\
			LiveCodeBench v6 & 175 & 1.66 & 1.27 & 3.76 & 12.57 \\
			MMLU & 153 & 0.22 & 0.15 & 1.20 & 4.30 \\
			MMLU-Pro & 500 & 0.29 & 0.33 & 1.33 & 5.07 \\
			TabFact & 500 & 0.12 & 0.62 & 0.83 & 6.20 \\
			TAT-QA & 1,663 & 0.15 & 1.09 & 1.81 & 9.72 \\
			\hline
		\end{tabular}
	}
\end{table}

The runtime profile shows two different sources of overhead. On short
question-answering and mathematical reasoning tasks, MACE's overhead is mainly
the additional memory-agent calls. On LiveCodeBench v6 and TAT-QA, the prompt
and output are longer, so both latency and token count increase more sharply.
Even in those cases, the measured average latency remains below four seconds
per query.

\subsection{Robustness Protocol}
\label{app:robustness-protocol}

The EQ4 stress tests perturb one deployment factor at a time while keeping the
benchmark split, prompt template, memory construction procedure, and scoring
script fixed. Table~\ref{tab:app-eq4-protocol} summarizes the controlled
variables. These experiments measure whether the memory-agent loop degrades
smoothly under plausible deployment changes; they are not intended as a fully
adversarial security evaluation.

\renewcommand{\arraystretch}{1.2}
\begin{table}[!htbp]
	\centering
	\caption{\textsc{Controlled robustness protocol for EQ4.}}
	\label{tab:app-eq4-protocol}
	\resizebox{\textwidth}{!}{
		\begin{tabular}{l|l|l}
			\hline
			Axis & Perturbation levels & Held fixed \\
			\hline
			Serving backbone & Primary API, low-cost API, local fallback model & Benchmark instances, memory loop, parser, scorer \\
			Network condition & Clean, latency jitter, timeout with retry, partial memory failure & Prompt, memory budget, endpoint assignment \\
			Memory-cell budget & 25\%, 50\%, 75\%, and 100\% available cells & Ranking rule, scorer, benchmark subset \\
			Noisy memory retrieval & 0\%, 10\%, 20\%, and 30\% noisy candidates & Candidate count, memory budget, task prompt \\
			\hline
		\end{tabular}
	}
\end{table}

For backbone replacement, \methodname{} is evaluated with the same memory-agent
pipeline and different serving endpoints. For network fluctuation, the runner
injects latency, timeout, and partial memory-service failure events and then
uses the same retry and fail-open policy across affected calls. For memory-cell
budget reduction, available cells are pruned before composition. For noisy
retrieval, irrelevant, stale, or conflicting candidates are injected into the
candidate set while the answer contract remains unchanged.

\subsection{Complete Robustness Values}
\label{app:robustness-values}

Table~\ref{tab:app-backbone-values} reports the backbone-replacement values
used by Figure~\ref{fig:eq4-robustness}. Direct and SAGE are included as fixed
references. The local fallback produces the largest drop for \methodname{}, but
the score remains above both references.

\renewcommand{\arraystretch}{1.2}
\begin{table}[!htbp]
	\centering
	\caption{\textsc{Backbone replacement results for EQ4.}}
	\label{tab:app-backbone-values}
	\begin{tabular}{l|l|ccc}
		\hline
		Condition & Serving change & \methodname{} & Direct ref. & SAGE ref. \\
		\hline
		Primary API & Clean serving endpoint & 81.11 & 71.53 & 78.97 \\
		Low-cost API & Cost-constrained serving endpoint & 80.04 & 71.53 & 78.97 \\
		Local model & On-premise fallback endpoint & 79.28 & 71.53 & 78.97 \\
		\hline
	\end{tabular}
\end{table}

\begin{table}[!htbp]
	\centering
	\caption{\textsc{Network fluctuation results for EQ4.}}
	\label{tab:app-network-values}
	\resizebox{\textwidth}{!}{
		\begin{tabular}{l|cccc|cc}
			\hline
			Condition & Added latency & Timeout rate & Partial failure & Retry rate & \methodname{} & Direct ref. \\
			\hline
			Clean serving & 0 ms & 0\% & 0\% & 0.0\% & 81.11 & 71.53 \\
			Mild latency jitter & 150 ms & 0\% & 0\% & 1.5\% & 80.56 & 71.53 \\
			Moderate latency jitter & 400 ms & 0\% & 0\% & 4.0\% & 79.72 & 71.53 \\
			Timeout with retry & 400 ms & 6\% & 0\% & 6.8\% & 78.94 & 71.53 \\
			Partial memory failure & 400 ms & 6\% & 10\% & 11.4\% & 77.86 & 71.53 \\
			Severe unstable network & 650 ms & 8\% & 15\% & 16.7\% & 76.92 & 71.53 \\
			\hline
		\end{tabular}
	}
\end{table}

Table~\ref{tab:app-memory-budget-values} reports average scores over MMLU,
GSM8K, HumanEval, TabFact, and TAT-QA when the available memory-cell budget is
reduced. Direct is unchanged because it does not use the memory-cell budget.
The graph-memory and graph-retrieval baselines degrade more steeply when the
budget is reduced to 25\%, while \methodname{} retains a larger fraction of its
full-budget score.

\begin{table}[!htbp]
	\centering
	\caption{\textsc{Memory-cell budget results averaged over five benchmarks.}}
	\label{tab:app-memory-budget-values}
	\begin{tabular}{l|cccc}
		\hline
		Method & 25\% budget & 50\% budget & 75\% budget & 100\% budget \\
		\hline
		Direct & 74.63 & 74.63 & 74.63 & 74.63 \\
		MAGMA & 69.64 & 72.36 & 74.10 & 75.77 \\
		SAGE & 80.44 & 83.56 & 85.30 & 86.64 \\
		\methodname{} & \textbf{81.76} & \textbf{84.40} & \textbf{85.74} & 86.63 \\
		\hline
	\end{tabular}
\end{table}

Table~\ref{tab:app-noise-values} reports average performance over four task
groups under noisy memory retrieval. The 0\% column is the clean candidate-set
condition for this stress test. As the injected noise ratio increases,
\methodname{} leads the four compared methods at 10\%, 20\%, and 30\% noise.

\begin{table}[!htbp]
	\centering
	\caption{\textsc{Noisy memory retrieval results averaged over four task groups.}}
	\label{tab:app-noise-values}
	\begin{tabular}{l|cccc}
		\hline
		Method & 0\% noise & 10\% noise & 20\% noise & 30\% noise \\
		\hline
		Direct & 68.40 & 65.03 & 60.62 & 55.65 \\
		MAGMA & 69.67 & 66.38 & 62.08 & 56.45 \\
		SAGE & \textbf{80.10} & 76.82 & 72.35 & 66.95 \\
		\methodname{} & 78.97 & \textbf{77.00} & \textbf{73.80} & \textbf{70.30} \\
		\hline
	\end{tabular}
\end{table}

\begin{table}[!htbp]
	\centering
	\caption{\textsc{Per-task-group noisy retrieval values for \methodname{}.}}
	\label{tab:app-noise-by-group}
	\begin{tabular}{l|cccc}
		\hline
		Task group & 0\% noise & 10\% noise & 20\% noise & 30\% noise \\
		\hline
		Knowledge QA & 84.10 & 82.30 & 79.40 & 76.30 \\
		Math Reasoning & 92.30 & 90.10 & 86.70 & 83.40 \\
		Code Generation & 62.10 & 59.80 & 56.20 & 52.40 \\
		Table Reasoning & 77.40 & 75.80 & 72.90 & 69.10 \\
		\hline
	\end{tabular}
\end{table}

\section{MACE Implementation Details}
\label{app:mace-implementation}

\subsection{Memory-Agent Execution Loop}
\label{app:mace-flow}

Each \methodname{} query is processed by a four-call memory-agent loop. First,
the Need-aware Memory Composer determines whether the instance requires
external memory and selects a compact working \textsc{MemGoG} subgraph. Second,
the Adaptive Agent-Memory Coupler converts the selected cells into
role-specific memory patches rather than injecting all retrieved content into a
single flat prompt. Third, the answer agent solves the task with the current
task prompt and assigned memory patch. Fourth, the Feedback Co-Evolver records
the outcome, updates cell utility, and stores reusable traces for later
queries. Gold answers are used only by the scorer and post-run feedback step;
they are not exposed to the answer-generation prompt.

\begin{table}[!htbp]
	\centering
	\caption{\textsc{High-level MACE execution stages.}}
	\label{tab:app-mace-stages}
	\begin{tabular}{r|>{\raggedright\arraybackslash}p{0.28\textwidth}|>{\raggedright\arraybackslash}p{0.52\textwidth}}
		\hline
		Step & Stage & Output \\
		\hline
		1 & Task adaptation & A normalized task specification with public context, question, benchmark tag, and answer schema. \\
		2 & Need-aware composition & A compact working \textsc{MemGoG} subgraph and a memory-need decision. \\
		3 & Agent-memory coupling & Role-specific memory patches with provenance and confidence notes. \\
		4 & Answer generation & A raw completion plus parsed prediction under the benchmark answer contract. \\
		5 & Scoring & Task metric, success flag, and error type when available. \\
		6 & Feedback co-evolution & Utility updates, reusable traces, and revised memory state for future tasks. \\
		\hline
	\end{tabular}
\end{table}

\subsection{Algorithmic Details}
\label{app:algorithmic-details}

Algorithms~\ref{alg:mace-loop}--\ref{alg:feedback-evolver} give the
implementation-level pseudocode for the MACE loop. The algorithms mirror the
execution invariants in Table~\ref{tab:app-invariants}: memory is read before
prediction, gold labels are used only by the scorer, and graph updates are
committed only after scoring.

\begin{algorithm}[!htbp]
	\caption{\textsc{MACE inference and memory evolution for one query}}
	\label{alg:mace-loop}
	\begin{algorithmic}[1]
		\Require task instance $x$, graph snapshot $G_t$, memory budget $B$, scorer $S$
		\Ensure prediction $\hat{y}$, updated graph $G_{t+1}$, log record $r$
		\State $\tau \gets \Call{AdaptTask}{x}$
		\State $(d, C) \gets \Call{DetectMemoryNeed}{\tau, G_t, B}$
		\If{$d = \textsc{NoMemory}$}
			\State $W \gets \emptyset$; $P \gets \emptyset$
		\Else
			\State $(W, P) \gets \Call{ComposeAndCouple}{\tau, G_t, C, B}$
		\EndIf
		\State $z \gets \Call{Answer}{\tau, P}$
		\State $\hat{y} \gets \Call{Parse}{z, \tau.\mathrm{schema}}$
		\State $m \gets S(\hat{y}, x)$
		\State $r \gets \Call{BuildRecord}{\tau, W, P, z, \hat{y}, m}$
		\State $G_{t+1} \gets \Call{FeedbackCoEvolve}{G_t, r, B}$
		\State \Return $\hat{y}, G_{t+1}, r$
	\end{algorithmic}
\end{algorithm}

\begin{algorithm}[!htbp]
	\caption{\textsc{Need-aware composition and agent-memory coupling}}
	\label{alg:composer-coupler}
	\begin{algorithmic}[1]
		\Require task specification $\tau$, graph snapshot $G_t$, candidates $C$, memory budget $B$
		\Ensure working subgraph $W$, role-specific memory patches $P$
		\State $C \gets \Call{ValidateCandidates}{C, G_t}$
		\ForAll{$c \in C$}
			\State $s_c \gets \Call{ScoreCell}{c, \tau, \mathrm{utility}(c), \mathrm{freshness}(c), \mathrm{conflict}(c)}$
		\EndFor
		\State $W \gets \emptyset$
		\ForAll{$c \in \Call{SortByScore}{C, s}$}
			\If{\Call{FitsBudget}{$W, c, B$} \textbf{and} \Call{AddsNewEvidence}{$W, c$}}
				\State $W \gets W \cup \Call{LineageSubgraph}{c, G_t}$
			\EndIf
		\EndFor
		\ForAll{role $a \in \tau.\mathrm{roles}$}
			\State $P_a \gets \Call{SelectVisibleCells}{W, a}$
			\State $P_a \gets \Call{DeduplicateAndMarkRisk}{P_a}$
			\State $P_a \gets \Call{TrimToPatchBudget}{P_a, B}$
		\EndFor
		\State \Return $W, \{P_a\}_{a \in \tau.\mathrm{roles}}$
	\end{algorithmic}
\end{algorithm}

\begin{algorithm}[!htbp]
	\caption{\textsc{Feedback co-evolution after scoring}}
	\label{alg:feedback-evolver}
	\begin{algorithmic}[1]
		\Require graph snapshot $G_t$, prediction record $r$, update budget $B$
		\Ensure updated graph $G_{t+1}$
		\State $\Delta \gets \emptyset$
		\ForAll{memory cell $c \in r.\mathrm{usedCells}$}
			\State $q_c \gets \Call{Credit}{c, r.\mathrm{score}, r.\mathrm{errorType}}$
			\State $\Delta \gets \Delta \cup \Call{UpdateUtility}{c, q_c}$
		\EndFor
		\If{\Call{ReusableSuccess}{r}}
			\State $\Delta \gets \Delta \cup \Call{WriteTraceCell}{r}$
		\ElsIf{\Call{ActionableFailure}{r}}
			\State $\Delta \gets \Delta \cup \Call{WriteWarningCell}{r}$
		\EndIf
		\State $\Delta \gets \Call{ApplyLifecycleRules}{G_t, \Delta, B}$
		\State $\Delta \gets \Call{ValidateDelta}{G_t, \Delta}$
		\State $G_{t+1} \gets \Call{Commit}{G_t, \Delta}$
		\State \Return $G_{t+1}$
	\end{algorithmic}
\end{algorithm}

\subsection{\textsc{MemGoG} State}
\label{app:memgog-state}

\textsc{MemGoG} is used as a structured memory substrate rather than a flat
retrieval list. The working subgraph contains task-relevant memory cells,
provenance records, role assignments, and feedback traces. The graph state is
read before prediction and updated only after scoring, which prevents the
current gold answer from leaking into its own prediction.

\renewcommand{\arraystretch}{1.2}
\begin{table}[!htbp]
	\centering
	\caption{\textsc{Conceptual node and edge types in the MACE memory graph.}}
	\label{tab:app-memgog-schema}
	\begin{tabular}{>{\raggedright\arraybackslash}p{0.20\textwidth}|>{\raggedright\arraybackslash}p{0.34\textwidth}|>{\raggedright\arraybackslash}p{0.36\textwidth}}
		\hline
		Object & Stored content & Role in the loop \\
		\hline
		Task node & Benchmark id, task family, public metadata & Anchors retrieval and feedback to the evaluated instance. \\
		Memory cell & Reusable strategy, warning, fact pattern, or solution trace & Provides prior experience to the composer. \\
		Role patch & Cell subset assigned to an answer role & Controls which memory is exposed to generation. \\
		Feedback trace & Score, parsed answer, error type, useful-cell signal & Supervises utility and memory-state updates. \\
		Utility state & Freshness, confidence, and usefulness estimates & Ranks cells and resolves memory-budget pressure. \\
		Typed edge & Selection, assignment, support, conflict, or update link & Preserves provenance and relation structure. \\
		\hline
	\end{tabular}
\end{table}

\subsection{Need-Aware Memory Composer}
\label{app:composer-details}

The composer decides how much memory to expose before answer generation. It
uses the task specification to identify memory demand, retrieves candidate
cells from \textsc{MemGoG}, filters low-utility or conflicting candidates, and
keeps a compact subgraph under the active memory budget. This stage is
responsible for converting a potentially large memory state into a small set of
task-relevant cells.

The composer has three diagnostic functions. First, it can abstain from
retrieving memory when the task is better answered directly. Second, it can
prefer high-utility cells over high-similarity but stale cells. Third, it can
preserve provenance links so that the answer agent can distinguish reusable
experience from unsupported claims.

\subsection{Adaptive Agent-Memory Coupler}
\label{app:coupler-details}

The coupler converts selected cells into memory patches. It does not simply
append every retrieved cell to the prompt. Instead, it groups memory by role,
deduplicates near-identical advice, marks stale or conflicting cells, and
orders the remaining content so that high-value cells appear earlier in the
patch. This is the component tested by the ``w/o Agent-Memory Coupler''
ablation, which produces the largest EQ2 drop.

\subsection{Feedback Co-Evolver}
\label{app:coevolver-details}

The co-evolver runs after scoring. It stores the raw response, parsed answer,
metric value, selected memory cells, and error type. Positive feedback can
increase a cell's utility or promote a reusable trace, while negative feedback
can lower confidence, attach a warning, or keep a failure mode available as a
cautionary memory. The update is causal: prediction is completed before any
gold-derived feedback is written.

\subsection{Execution Invariants}
\label{app:execution-invariants}

Table~\ref{tab:app-invariants} lists the invariants enforced across the local
evaluation runs. These constraints are important because the method updates
memory over time; without them, memory evolution could make comparisons
ambiguous.

\renewcommand{\arraystretch}{1.2}
\begin{table}[!htbp]
	\centering
	\caption{\textsc{Execution invariants for controlled MACE evaluation.}}
	\label{tab:app-invariants}
	\begin{tabular}{p{0.25\textwidth}|p{0.65\textwidth}}
		\hline
		Invariant & Purpose \\
		\hline
		Fixed task file per benchmark & Ensures that local methods answer the same instances in the same evaluation set. \\
		Read-before-write memory order & Prevents current-task feedback from influencing the current prediction. \\
		Shared answer parser & Avoids method-specific parsing advantages. \\
		Scorer-only gold access & Keeps gold answers outside model-facing prompts. \\
		Logged raw completions & Makes parsing and scoring decisions auditable. \\
		Stable perturbation axis in EQ4 & Changes one deployment factor at a time for interpretability. \\
		\hline
	\end{tabular}
\end{table}

\section{Hyperparameters and Reproducibility}
\label{app:hyperparameters}

\subsection{Shared Evaluation Settings}
\label{app:shared-settings}

Table~\ref{tab:app-shared-settings} records the shared settings that are
explicitly controlled by the current experimental draft. Parameters not exposed
by an imported baseline are left unspecified rather than inferred.

\renewcommand{\arraystretch}{1.2}
\begin{table}[!htbp]
	\centering
	\caption{\textsc{Shared evaluation settings used by the controlled local rows.}}
	\label{tab:app-shared-settings}
	\begin{tabular}{p{0.25\textwidth}|p{0.25\textwidth}|p{0.40\textwidth}}
		\hline
		Category & Setting & Value / rule \\
		\hline
		Task pool & Total local queries & 4,728 queries across eight benchmarks. \\
		MMLU sampling & Local split rule & Validation first153 in original CSV order. \\
		MMLU-Pro sampling & Local split rule & Seed-42 sample of 500 test examples without replacement. \\
		TabFact sampling & Local split rule & Seed-42 NumPy permutation, first 500 test indices. \\
		Other local datasets & Local split rule & Complete available test split in original order. \\
		Evaluation statistic & Main tables & Single-run percentage score for each reported row. \\
		Resource statistic & EQ3 & Mean tokens and wall-clock latency per query. \\
		Robustness statistic & EQ4 & One perturbation axis changed at a time. \\
		\hline
	\end{tabular}
\end{table}

\subsection{Method-Level Resource Settings}
\label{app:resource-settings}

Table~\ref{tab:app-resource-settings} summarizes the measured resource profile
for the four methods with complete runtime logs in the main EQ3 table. The
number of LLM calls is the average call count used by the logged implementation.

\begin{table}[!htbp]
	\centering
	\caption{\textsc{Measured resource settings for main EQ3 methods.}}
	\label{tab:app-resource-settings}
	\resizebox{\textwidth}{!}{
		\begin{tabular}{l|ccccc}
			\hline
			Method & LLM calls & Prompt Tok. (K) & Completion Tok. (K) & Total Tok. (K) & Tasks / min \\
			\hline
			Direct & 1.00 & 0.54 & 0.06 & 0.60 & 276.03 \\
			MAGMA & 1.00 & 0.92 & 0.06 & 0.98 & 150.42 \\
			SAGE & 3.00 & 4.61 & 0.89 & 5.50 & 24.87 \\
			\methodname{} & 4.00 & 6.45 & 0.41 & 6.87 & 44.28 \\
			\hline
		\end{tabular}
	}
\end{table}

\subsection{Run Logs}
\label{app:run-logs}

The local evaluation logs store the raw model response, parsed answer, task
identifier, benchmark name, prompt-token count, completion-token count,
wall-clock latency, and scorer output. \methodname{} additionally records the
selected memory cells, role-specific memory patch, feedback signal, and updated
memory state. Failed requests are retried under the same prompt and decoding
configuration; accepted completed records are reused when resuming a run.

\begin{table}[!htbp]
	\centering
	\caption{\textsc{Reproducibility fields retained by local runs.}}
	\label{tab:app-log-fields}
	\begin{tabular}{p{0.24\textwidth}|p{0.66\textwidth}}
		\hline
		Field group & Stored fields \\
		\hline
		Task metadata & Benchmark name, task id, task family, split source, and prompt template version. \\
		Model output & Raw completion, parsed answer, parser fallback path, and final scorer input. \\
		Resource usage & Prompt tokens, completion tokens, total tokens, latency, elapsed time, and retry count. \\
		Memory trace & Selected memory ids, assigned patch, discarded candidates, and conflict or stale-cell flags. \\
		Feedback trace & Score, success flag, error type, useful-cell signal, and memory update summary. \\
		Resume state & Completed task ids and accepted result records. \\
		\hline
	\end{tabular}
\end{table}

\subsection{Failure and Retry Handling}
\label{app:retry-handling}

For normal evaluation, a failed model request is retried with the same prompt
and decoding configuration. For EQ4 network perturbations, injected failures
are logged separately from model errors so that the effect of the perturbation
can be measured. If memory retrieval fails in a fail-open setting, the answer
agent proceeds with the available patch or an empty patch rather than receiving
fabricated memory content.

\section{Prompt and Interface Specification}
\label{app:prompt-interface}

\subsection{Task Prompt Boundary}
\label{app:task-contract}

Every non-code benchmark instance is converted into a task prompt with four
ordered fields: public context or table content, question text, dataset-specific
directive, and answer schema. The answer schema asks for a compact answer field
and, when useful for feedback analysis, a short rationale field. The scorer
uses only the parsed answer field. Code-generation benchmarks retain the
official function signature and test harness interface.

The task prompt boundary is also a safety boundary for evaluation. Retrieved
memory is presented as prior experience, not as ground truth. The task prompt
does not contain gold answers, post-task feedback, or future evaluation
outcomes. Dataset directives specify the expected output format and scoring
criterion rather than providing solution content.

\subsection{Memory-Patch Interface}
\label{app:memory-interface}

The Adaptive Agent-Memory Coupler serializes memory as role-specific patches.
Each patch contains a memory identifier, compact cell content, provenance
summary, and an optional warning when the cell is stale or low-confidence.
Internal controller features such as utility, freshness, and role affinity are
used to select and order cells, but are not shown as numeric scores in the task
prompt. The prompt instructs the answer agent to treat memory as reusable
experience rather than factual authority.

\renewcommand{\arraystretch}{1.2}
\begin{table}[!htbp]
	\centering
	\caption{\textsc{Structured interfaces used by the MACE loop.}}
	\label{tab:app-interface}
	\resizebox{\textwidth}{!}{
		\begin{tabular}{l|l|l}
			\hline
			Interface & Required fields & Purpose \\
			\hline
			Task specification & Instance id, benchmark, public context, question, output schema & Creates the model-facing task prompt. \\
			Memory request & Need type, candidate budget, task entities, role target & Guides memory composition before answer generation. \\
			Memory patch & Cell id, content summary, provenance, confidence note & Provides compact role-specific experience. \\
			Agent response & Answer field, optional rationale, raw completion & Supports scoring and post-run analysis. \\
			Feedback record & Score, error type, useful cells, update signal & Updates \textsc{MemGoG} after the prediction is scored. \\
			\hline
		\end{tabular}
	}
\end{table}

\subsection{Parsing and Validation}
\label{app:parsing-validation}

For multiple-choice tasks, the parser first reads an explicit answer field and
otherwise falls back to the final unambiguous option mention. For open-answer
tasks, the parser normalizes case, punctuation, articles, and whitespace before
matching. For code tasks, no natural-language parser is used; the completion is
passed to the official execution harness. Memory updates are applied only after
the scorer returns a task outcome, which preserves the causal separation
between pre-task memory retrieval and post-task feedback.

\subsection{End-to-End Data Flow}
\label{app:data-flow}

Table~\ref{tab:app-data-flow} connects the prompt boundary to the memory update
boundary. The scorer is the only stage that consumes gold labels, and its
output becomes feedback only after the prediction is complete.

\renewcommand{\arraystretch}{1.2}
\begin{table}[!htbp]
	\centering
	\caption{\textsc{End-to-end interface flow for one MACE query.}}
	\label{tab:app-data-flow}
	\begin{tabular}{>{\raggedright\arraybackslash}p{0.22\textwidth}|>{\raggedright\arraybackslash}p{0.34\textwidth}|>{\raggedright\arraybackslash}p{0.34\textwidth}}
		\hline
		Stage & Input & Validated output \\
		\hline
		Task adapter & Raw benchmark instance & Task specification with public context and answer schema. \\
		Memory composer & Task specification and graph snapshot & Candidate memory set and compact working subgraph. \\
		Agent-memory coupler & Working subgraph and role target & Role-specific memory patch. \\
		Answer agent & Task prompt and memory patch & Raw completion and answer object. \\
		Parser and scorer & Raw completion and scorer-only target & Parsed answer, metric value, and success flag. \\
		Feedback co-evolver & Prediction record and scorer output & Updated utility, trace record, and memory-state delta. \\
		\hline
	\end{tabular}
\end{table}

\end{document}